\documentclass{article}

\PassOptionsToPackage{numbers, compress}{natbib}
     \usepackage[preprint]{neurips_2019}

\usepackage[utf8]{inputenc} 
\usepackage[T1]{fontenc}    
\usepackage{hyperref}       
\usepackage{url}            
\usepackage{booktabs}       
\usepackage{amsfonts}       
\usepackage{nicefrac}       
\usepackage{microtype}      

\usepackage{graphicx}
\usepackage{subcaption}
\usepackage[ruled,vlined,noresetcount]{algorithm2e} 

\SetKwComment{Comment}{$\triangleright$\ }{}

\title{Snooping Attacks on Deep Reinforcement Learning}

\author{%
  Matthew Inkawhich\\
  Duke University\\
  \texttt{mji5@duke.edu} \\
  \And
  Yiran Chen\\
  Duke University\\
  \texttt{yiran.chen@duke.edu} \\
  \And
  Hai Li \\
  Duke University \\
  \texttt{hai.li@duke.edu} \\
}

\begin{document}
\maketitle

\begin{abstract}  
Adversarial attacks have exposed a significant security vulnerability in state-of-the-art machine learning models. Among these models include deep reinforcement learning agents. The existing methods for attacking reinforcement learning agents assume the adversary either has access to the target agent's learned parameters or the environment that the agent interacts with. In this work, we propose a new class of threat models, called snooping threat models, that are unique to reinforcement learning. In these snooping threat models, the adversary does not have the ability to interact with the target agent's environment, and can only eavesdrop on the action and reward signals being exchanged between agent and environment. We show that adversaries operating in these highly constrained threat models can still launch devastating attacks against the target agent by training proxy models on related tasks and leveraging the transferability of adversarial examples.
\end{abstract}

\section{Introduction}
The deep learning revolution has put neural networks at the forefront of the machine learning research and production landscape. Recently, it has been shown that deep neural networks (DNNs) are effective function approximators for solving complex reinforcement learning (RL) problems. \citet{Mnih2013PlayingAW} demonstrated that we can leverage the feature extraction capabilities of convolutional neural networks to enable RL agents to learn to play Atari games from raw pixels. Since then, there has been an influx of work dedicated to the field of deep RL (DRL), enabling powerful new solutions for tasks such as game playing \citep{Mnih2013PlayingAW, Mnih2016AsynchronousMF, Silver2016MasteringTG, Schulman2015TrustRP}, continuous control \citep{Lillicrap2016ContinuousCW, Schulman2017ProximalPO, Silver2014DeterministicPG}, complex robot manipulation \citep{Levine2016EndtoEndTO, Pinto2018AsymmetricAC, Pathak2018ZeroShotVI}, and even autonomous vehicle operation \citep{Bojarski2016EndTE, Sallab2017DeepRL}.

Despite the numerous successes of the applications of deep neural networks, these models have a significant Achilles' heel. Studies show that these largely uninterpretable models are vulnerable to small adversarial perturbations to the input that cause the models to perform poorly \citep{Szegedy2014IntriguingPO, Goodfellow2015ExplainingAH}. Recent work has uncovered a variety of adversarial attack algorithms in the domain of image classification \citep{Goodfellow2015ExplainingAH, Carlini2017TowardsET, Papernot2016TheLO, Dong2018BoostingAA}. Alarmingly, results show that adversarial examples effective against one model are also effective against other models trained on the same task \citep{Goodfellow2015ExplainingAH}. This transferability property has enabled attacks against models in black-box settings \citep{Papernot2017PracticalBA, Inkawhich2019FeatureSpaceP}.

As of now, adversarial attacks against DRL agents primarily consider the white-box threat model, in which the adversary is assumed to have access to the target model's architecture and parameters. To our knowledge, the study of black-box attacks against DRL has been limited to the acknowledgement of the transferability of adversarial examples across policies and  algorithms \citep{Huang2017AdversarialAO}. Thus, the current method for black-box attacks against DRL agents involves training a surrogate agent, which makes the rather unrealistic assumption that the adversary has unbounded access to the environment. In general, the black-box threat model in RL is more complex than it is in supervised learning. First, in the data generating process for supervised learning the model draws i.i.d. samples from a static dataset. In RL, data generation is in the form of an environment from which new data samples are conditioned on the current environment state and the agent's actions. This means that an RL adversary cannot trivially approximate the training data generating process the way a supervised learning adversary can. Also, the RL problem consists of an exchange of three signals between the environment and agent (state, action, reward). Even if an RL adversary seeks to approximate the agent or environment using some other method (e.g., inverse RL), they would need access to all signals.

In this work, we investigate threat models for DRL under the assumption that the adversary does not have access to the environment that the target agent interacts with. Specifically, we explore the potential effectiveness of an adversary that only has the ability to eavesdrop on a subset of the RL signals at each time step. We dub these the snooping threat models. For a real-world example of the snooping threat models of a vision-based DRL system, consider a camera-equipped mobile system such as an autonomous vehicle. These systems typically send the camera feed via a wireless network to a server running an instance of the trained DRL agent, which processes the images and decides how to act in the given state. As with all current adversarial threat models, the adversary is assumed to have the ability to alter the input of the model. Action signals are transmitted from the server back to the drone via a separate channel. Finally, reward signals may come from a variety of sources. Given this scenario, it is entirely possible that the adversary cannot access or eavesdrop on all signals due to private channels, encryption, etc.

We show that by training proxy models on tasks \textit{similar to} the target agent's task, we are able to craft adversarial inputs that significantly reduce the performance of target agents trained on various Atari environments. We feel this to be a more practical threat model for real-world systems. Overall, the contributions of this paper can be summarized as follows:
\begin{itemize}
  \item We define the complex threat model taxonomy for snooping attacks on DRL agents.
  \item We propose the use of proxy models for launching effective adversarial attacks under the different snooping threat models.
  \item We conduct extensive experiments using the attacks against the state-of-the-art DRL algorithms DQN \citep{Mnih2013PlayingAW} and PPO \citep{Schulman2017ProximalPO}.
  \item We empirically show that adversarial examples transfer between models that are trained with very different objectives as long as the tasks are related.
\end{itemize}

\section{Related work}
Adversarial machine learning is a research area that investigates the vulnerabilities of deep neural networks to small adversarial perturbations. In this work, we focus on evasion attacks, which are test-time attacks designed to fool a trained model. This topic was popularized in the image classification domain in the white-box setting, where \citet{Szegedy2014IntriguingPO} discovered that adding imperceptible noise to an image causes the model to predict the incorrect class with high confidence. \citet{Goodfellow2015ExplainingAH} followed by suggesting a fast gradient method for crafting adversarial examples in an inexpensive way by assuming local linearity in the target model. \citet{Goodfellow2015ExplainingAH} also observed that adversarial examples crafted using the gradients of one model transfer to other models trained on the same task. \citet{Papernot2017PracticalBA} leveraged this concept of transferability and introduced a framework for black-box attacks that uses the target model as an oracle to train a surrogate model with a similar decision surface, and uses the surrogate to craft adversarial examples that also fool the target.

In the DRL domain, \citet{Huang2017AdversarialAO} showed that the fast gradient method \citep{Goodfellow2015ExplainingAH} can be extended to fool agents on the Atari benchmark. \citet{Lin2017TacticsOA} introduced a strategically-timed attack, which seeks to make perturbations sparse in time to render them more difficult to detect. \citet{Lin2017TacticsOA} further presented an enchanting attack that strives to lure the target agent into a specified state. \citet{Behzadan2017VulnerabilityOD} demonstrated that adversarial examples can be used during training to corrupt the learning process of an agent on the Atari Pong environment. Adversarial examples have also been used for hardening DRL models to environmental parameter variations in continuous control environments \citep{Pattanaik2018RobustDR}. Overall, adversarial attacks on DRL agents have primarily been limited to the white-box setting.

\section{Preliminaries}
\subsection{Adversarial example crafting}
The goal of adversarial example crafting is to apply an imperceptible perturbation to a benign input such that the perturbed input fools the target model. In this work, we consider variants of the Fast Gradient Method (FGM) \citep{Goodfellow2015ExplainingAH} because it is computationally efficient and generalizes to models other than classifiers. FGM is a one-step method that assumes linearity of the decision surface around a given sample. With this assumption, the optimal perturbation for an input $x$ is in the direction to maximize the loss $J$. We consider constraining the perturbations under the $L_{\infty}$ and $L_2$ norm bounds.

\begin{figure}[h]
    \centering
    \includegraphics[width=.7\columnwidth]{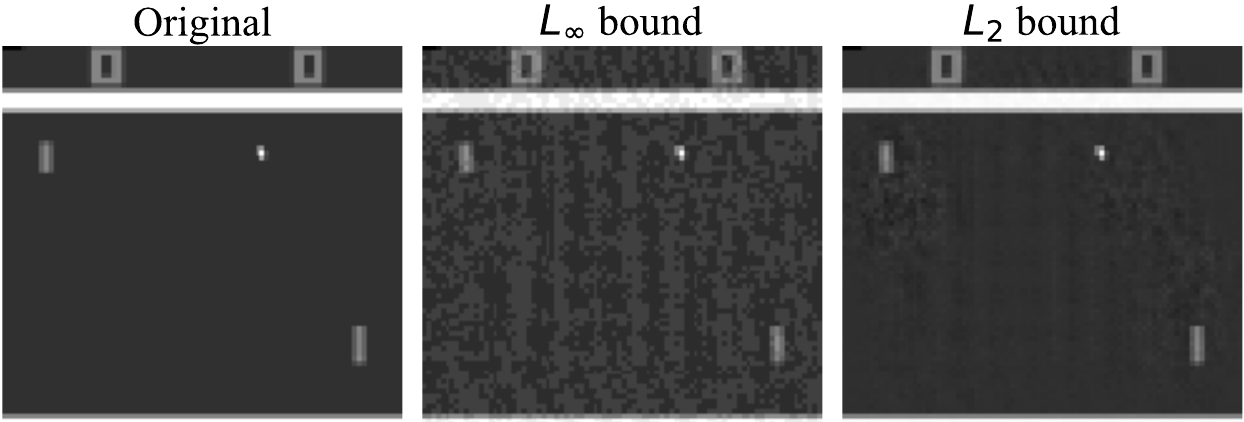}
    \caption{Adversarial example comparison between FGM attacks under different norm constraints on the Pong environment. The $L_\infty$ bounded example was crafted with $\epsilon = 0.03$ and the $L_2$ bounded example used $\epsilon = 2.4$. These examples were crafted using the psychic proxy model (see \ref{ss:s}).}
    \label{fig:example_compare}
\end{figure}

Under the $L_{\infty}$ norm bound $\left\Vert x^*-x \right\Vert_{\infty} \leq \epsilon$, an adversarial example $x^*$ is generated as
\begin{equation} \label{eq:Linf}
x^* = x + \epsilon*sign(\nabla_{x}J(x,y))
\end{equation}
where $y$ is the label. Under the $L_2$ norm bound $\left\Vert x^*-x \right\Vert_{2} \leq \epsilon$, $x^*$ is generated as
\begin{equation} \label{eq:L2}
x^* = x + \epsilon*\frac{\nabla_{x}J(x,y)}{\left\Vert \nabla_{x}J(x,y) \right\Vert_{2}} \mathrm{.}
\end{equation}
We also consider a momentum iterative variant of FGM (MIFGM) introduced by \citet{Dong2018BoostingAA}. This variant disregards the linearity assumption and iteratively perturbs while accumulating a velocity vector in the direction of the gradient to stabilize the perturbation direction and combat overfitting. The velocity vector $g$ is accumulated as 
\begin{equation} \label{eq:mifgsm}
g_{t+1} = \mu * g_t + \frac{\nabla_{x}J(x^*_t,y)}{\left\Vert \nabla_{x}J(x^*_t,y) \right\Vert_{1}} \mathrm{.}
\end{equation}
To craft the example, at each iteration we use Equations \ref{eq:Linf} and \ref{eq:L2}, but substitute $\nabla_{x}J(x,y)$ with $g_{t+1}$. A visualization of $L_{\infty}$ and $L_2$ constrained FGM perturbations on Pong are shown in Figure \ref{fig:example_compare}.

\subsection{Reinforcement learning}
Reinforcement Learning is a machine learning paradigm based on sequential interactions between an agent and an environment in which the agent attempts to learn a policy $\pi$ to maximize a total reward \citep{Sutton1988ReinforcementLA}. This interaction is formalized as a Markov Decision Process (MDP), which is the tuple $\langle \mathcal{S}, \mathcal{A}, \mathcal{P}, \mathcal{R}, \gamma \rangle$. $\mathcal{S}$ is a finite set of states, $\mathcal{A}$ is a finite set of actions, $\mathcal{P}(s,a,s') = P(S_{t+1} = s' | S_t = s , A_t = a)$ is the state transition function, $\mathcal{R}(s,a)= \mathbb{E}[R_{t+1} | S_t = s, A_t = a]$ is the reward function and $\gamma \in [0,1]$ is a discount factor. The interaction can be described as follows: At each time step $t$, the environment sends a state $s_t$ to the agent. The agent decides on an action $a_t$ and dispatches it back to the environment. The environment responds by sending a reward $r_{t+1}$ to the agent along with the next state $s_{t+1}$, and the process repeats. In this work, we consider the state-of-the-art value-based DRL method DQN \citep{Mnih2013PlayingAW} and policy-based PPO \citep{Schulman2017ProximalPO}.

\subsubsection{Deep Q-Networks}
Q-learning is a value-based RL algorithm that estimates the cumulative discounted reward of each state-action pair, and chooses the agent's policy based on these estimated returns. In Q-learning, we approximate the function $Q^*(s,a)$, which gives the cumulative discounted reward for taking action $a$ in state $s$ and following an optimal policy $\pi^*$ thereafter \citep{Watkins1992TechnicalNQ}. To train, we iteratively optimize with a temporal difference loss $\delta$ based on the Bellman equation \citep{Bellman1966DynamicP}:
\begin{equation} \label{eq:tdloss}
\delta = Q(s,a) - (r + \gamma \max_a Q(s',a)) \mathrm{.}
\end{equation}
Once we have an acceptable approximation of $Q^*$, a common policy is to act $\epsilon$-greedily to aid in exploration. In an effort to apply Q-learning to problems with large state spaces such as Atari games, \citet{Mnih2013PlayingAW} combine convolutional neural networks with Q-learning to create deep Q-learning. In practice, this requires the use of a replay memory buffer to enable off-policy learning and a target network for more stable updates.

\subsubsection{Proximal policy optimization}
PPO \citep{Schulman2017ProximalPO} is a policy-based RL method, meaning that it directly approximates the policy $\pi_\theta(a|s)$ with a neural network with parameters $\theta$. The algorithm is typically implemented in an actor-critic framework which simultaneously learns to approximate $\pi_\theta(a|s)$ and a variance-reduced advantage estimate to stabilize the policy gradient. At each training step, PPO alternates between (1) collecting experience by running the current policy for a set number of time steps, (2) computing empirical returns and advantages, and (3) using batch learning to optimize a clipped surrogate objective that constrains the amount the updated policy can differ from the old policy:
\begin{equation} \label{eq:ppoloss}
L^{CLIP}(\theta) = \mathbb{E}_t \big[\min(r_t(\theta)\hat{A}_t, \mathrm{clip}(r_t(\theta), 1 - \alpha, 1 + \alpha)\hat{A}_t) \big]
\end{equation}
where $\alpha$ is a hyperparameter, $\hat{A}_t$ is the empirical advantage, and $r_t(\theta)$ is the probability ratio
\begin{equation} \label{eq:pporatio}
r_t(\theta) = \frac{\pi_{\theta}(a_t, s_t)}{\pi_{\theta_{old}}(a_t, s_t)} \mathrm{.}
\end{equation}

\begin{figure}[t]
    \centering
    \includegraphics[width=.8\columnwidth]{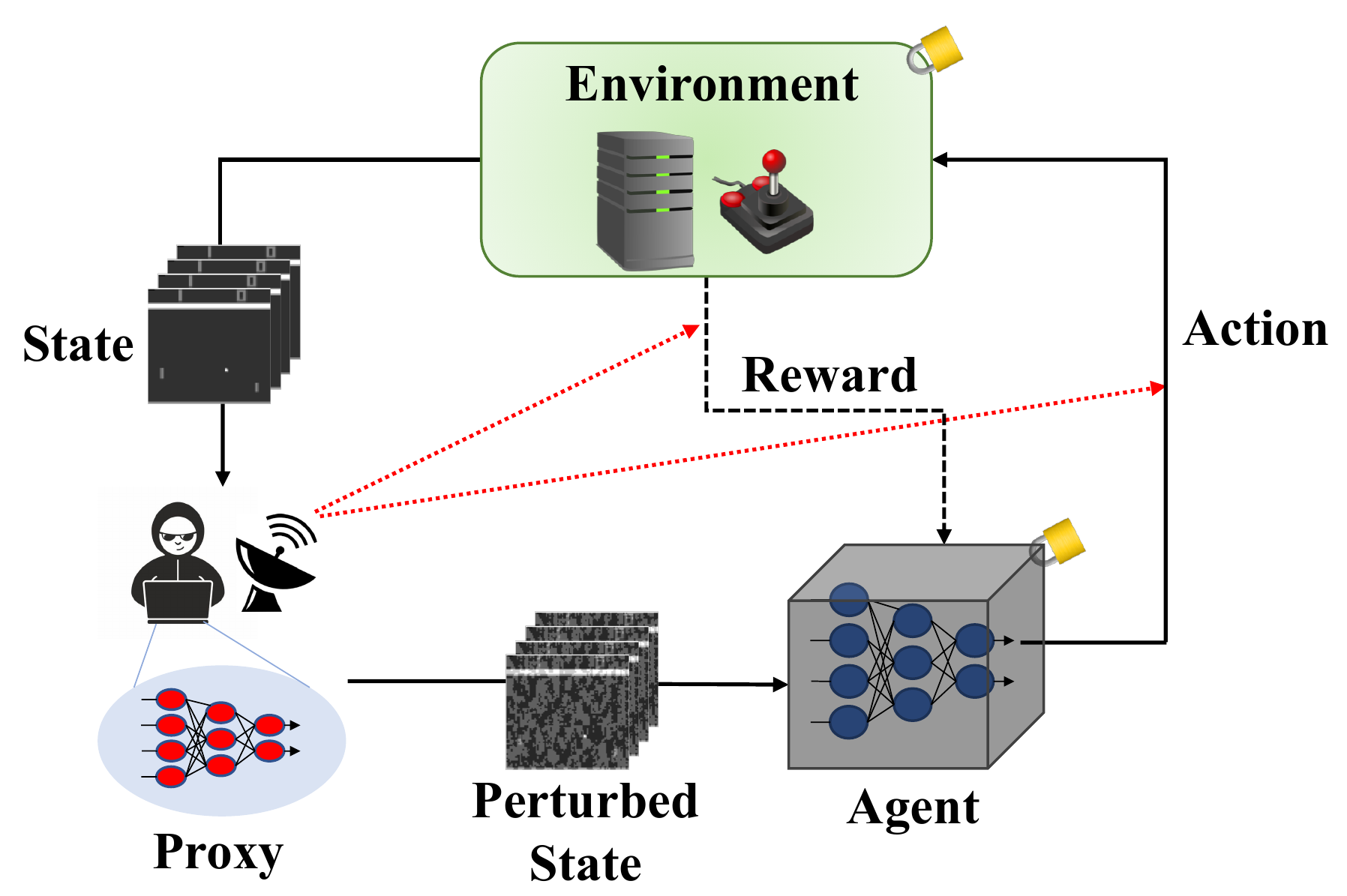}
    \caption{Snooping threat model overview for DRL.}
    \label{fig:main}
\end{figure}

\section{Snooping threat models}
A high level view of the snooping threat model for DRL is portrayed in Figure \ref{fig:main}. As in any threat model, we assume the adversary can intercept and manipulate the input game frames (i.e., the state). In the snooping family of threat models, the adversary does not have direct access to the environment, so training a surrogate DRL model and transferring examples from the surrogate to the target is ruled out. Instead, the adversary can only "snoop", or eavesdrop, on the natural interactions between the agent and environment. We define the \textbf{SRA} threat model to represent the scenario in which the adversary can snoop on reward and action signals, \textbf{SA} for when the adversary can only snoop on the actions, \textbf{SR} for when the adversary can only snoop on the rewards, and \textbf{S} for when the adversary only has access to the states. When the reward signal is hidden, the adversary is assumed to have no inclination about the goal of the agent. In the general case of game environments, this means that the adversary cannot explicitly infer the difference between winning and losing. When the action signal is hidden, not only is the adversary unable to determine which actions correspond to certain changes in the state space, but there is also no way of explicitly knowing the scope and magnitude of the action space at all.

\section{Attack strategies} \label{sec:methodology}

\subsection{Methodology} \label{ss:methodology}
Our general attack methodology is to leverage the transferability of adversarial examples. While this concept is not fully understood, recent efforts proclaim that transferability is primarily enabled by decision boundary similarity between models \citep{Tramr2017TheSO, Szegedy2014IntriguingPO, Liu2017DelvingIT}. We argue that this condition, however, generally overestimates the resources necessary to craft effective transferable adversarial examples. The snooping threat models for RL provides one such counterexample, in which the adversary cannot train a surrogate DRL model that closely resembles the target agent. Instead, we posit that if the adversary can train a proxy model that learns a task that is \textit{related} to the target agent's policy to maximize reward, adversarial examples crafted to fool the proxy will also fool the agent. This implies that even proxy models with radically different decision surface dimensionality than the target can reliably produce transferable attacks. We hypothesize that by learning a related task, the proxy will learn to extract and use input features in a similar fashion to the target agent, which is the only necessary condition for crafting transferable attacks. Our explicit goal is to train a proxy model $\mathcal{M}$ under a constraint subset $\kappa \in \{S, SR, SA, SRA\}$ parameterized by $\theta_{\mathcal{M}_{\kappa}}$ with loss $J_{\mathcal{M}_{\kappa}}$ that extracts similar input features and learns comparable implications of these features to the target agent $\mathcal{T}$ parameterized by $\theta_{\mathcal{T}}$ trained with loss $J_{\mathcal{T}}$ (e.g., temporal difference error). Using input gradients as an indication of feature saliency, our intention is to train a constrained proxy $\mathcal{M}_{\kappa}$ to optimize:
\begin{equation} \label{eq:main}
\arg\min_{\mathcal{M_{\kappa}}}\bigg[\frac{1}{N} \sum_{i = 1}^{N} \left \| \nabla_x J_{\mathcal{M}_{\kappa}}(\theta_{\mathcal{M}_{\kappa}}, x_i, y_i) - c \nabla_x J_{\mathcal{T}}(\theta_{\mathcal{T}}, x_i, z_i)\right \| \bigg]
\end{equation}
where $N$ is an arbitrary number of inputs,  $x_i$ is an input, $y_i$ is the true label for the proxy, $z_i$ is the pseudo-true label \citep{Huang2017AdversarialAO} for the agent, and $c$ is an unknown constant. We include $c$ as a scaling constant as we desire $\mathcal{M}_{\kappa}$ to produce similarly \textit{shaped} input gradients to $\mathcal{T}$'s, meaning the magnitude of the gradients do not necessarily have to align exactly. Note that this objective is a theoretical representation, and does not represent a tractable optimization objective. The exact choice of task and proxy $\mathcal{M}_{\kappa}$ is determined by our discretion.

Also, note that since we are attacking RL agents during test time, we will not have access to ground truth labels required by the FGM method (i.e. the $y$ in Equations \ref{eq:Linf} and \ref{eq:L2}), as RL data is inherently unlabeled. To work around this, we follow \citep{Huang2017AdversarialAO} and assume that our surrogates and proxies have converged to a reasonable minima, so we use their outputs as "truth". For surrogate agents or classifier proxies, we use the one-hot output of the model as truth, and for regressor proxies, we use the output (plus a small constant)\footnote{We must add a small constant to avoid computing a loss of zero.} as truth.

\subsection{S threat model} \label{ss:s}
The \textbf{S} threat model is the weakest snooping threat model. The adversary only has access to a stream of states, and has no inclination about the motivations or capabilities of the agent. In this case, we propose a proxy $\mathcal{M}_{S}$ that models the environment dynamics. Previous work shows that we can perform accurate next-frame prediction using the current state, action, and an $L_2$ reconstruction loss as long as $\mathcal{P}$ is roughly deterministic \citep{Oh2015ActionConditionalVP, Wang2017DeepAC, Leibfried2017ADL}. This model directly approximates the state transition distribution $\mathcal{P}(s,a,s')$. In \textbf{S} however, we do not have access to $\mathcal{A}$, so the best we can do is approximate an expectation of $\mathcal{P}$ under the target agent's policy $\pi_T$ using a \textit{psychic} proxy model:
\begin{equation} \label{eq:psychic}
psychic(s_t, \theta_P) \approx \mathbb{E}_{\pi_T}[P(s_{t+1} | s_t)] = \mathbb{E}_{a_t \sim \pi_T}[P(s_{t+1} | s_t, a_t)].
\end{equation}
Because the psychic learns the dynamics $\mathcal{P}$ under $\pi_T$, it effectively learns an encoding of $\pi_T$ in the state space. In other words, the displacement of the agent's representation in the state space (e.g. the pixels of the paddle in Breakout) from $s_t$ to $s_{t+1}$ inherently encodes a representation of the action $a_t$ taken and therefore an instance of the agent's policy $\pi_T(s_t)$. This approximation will be noisy in general due to displacements in the state space that are not the consequence of the agent's actions (e.g. the movement of the fish in Seaquest).
Despite the noise, the key is that a subset of the features learned by the psychic will be similar to the features learned by the target agent.

\subsection{SR threat model}
In the \textbf{SR} threat model, we can intercept the states and snoop on the rewards. Because we do not have access to the actions, we again cannot make any presumptions about the policy $\pi_T$ directly. However, by observing the rewards resulting from each state in a rollout we can get insight into the agent's motivations by training an \textit{assessor} proxy to estimate the value $V$ of a given state under $\pi_T$. Since the reward signal is what drives the policy $\pi_T$, the features learned to approximate $V^{\pi_T}$ must correspond to the features used by the agent itself.
\begin{equation} \label{eq:assessor}
assessor(s_t, \theta_A) \approx \mathbb{E}_{\pi_T}\Big[\sum\nolimits_{k=0}^\infty \gamma_t^{(k)}r_{t+k+1}\Big] = V^{\pi_T}(s_t)
\end{equation}
In our study we use empirical return to estimate value, meaning that we observe the immediate rewards at each step, and when the episode is over we retroactively compute and standardize the discounted sum of rewards at each state. During training, the assessor minimizes a Huber regression objective \citep{Girshick2015FastR} between predicted value and empirical value. Despite the fact that the $J_{\mathcal{M}_{SR}}$ objective is different than $J_{\mathcal{T}}$, the task of approximating $V^{\pi_T}$ is directly related to $\pi_T$. For example, value-based agents directly approximate a value estimate $Q^{\pi_T}(s_t, \pi_T(s_t)) = V^{\pi_T}(s_t)$, and for policy-based agents the explicit goal of the resulting policy is to maximize return, so implicitly the agent learns to assess the value of states.

\subsection{SA threat model}
In the \textbf{SA} threat model, we have access to a stream of states and can snoop on the actions taken by the agent at each step. Therefore, approximating the target's policy $\pi_T$ can be done directly with supervised learning. We can train an \textit{imitator} model to predict the action that the target will take at a given state as:
\begin{equation} \label{eq:imitator}
imitator(s_t, \theta_I) \approx \pi_T(s_t).
\end{equation}
Since the imitator is a classifier, we train it with cross-entropy loss ($J_{\mathcal{M}_{SA}}$) using the agent's actions as labels. Although the imitator is trained with a different objective, it is directly related to the target agent's goal. For value-based agents, the policy $\pi$ is an $\epsilon$-greedy extension of the learned Q-function and for policy-based agents, the function $\pi_T$ is directly approximated. This methodology is homologous to the black-box attack strategy in \citep{Papernot2017PracticalBA}, in which a substitute model is trained using the target model as an oracle. Note that if we were to have a continuous action space, the imitator would simply become a regression model.

\subsection{SRA threat model}
When we have the ability to eavesdrop on rewards \textit{and} actions, we can attack with an imitator, assessor, or psychic. However, with this additional information we can improve our attacks by strategically timing the perturbations to make them less detectable. \citet{Lin2017TacticsOA} show that we can accomplish this in a white-box setting by defining a preference function 
\begin{equation} \label{eq:strategicpreference}
c(s_t) = \max_{a_t}[Softmax(Q(s_t,a_t))]-\min_{a_t}[Softmax(Q(s_t,a_t))]
\end{equation}
and attacking when $c(s) \ge \beta$ for some threshold $\beta$. At first glance, it may seem possible to use a similar preference-based approach with our imitator or assessor models. However, these proxies are only ever exposed to one particular policy $\pi_T$ during training, so they lack the ability to assign credit to certain states or actions from the delayed reward signal. The RL agent, by contrast, learns this credit assignment through repeated trial and error during training. Nevertheless, we can approximate this ability using a combination of an assessor and an action-conditioned psychic (\textit{AC-psychic}), which predicts $s_{t+1}$ given $s_t$ and $a_t$ \citep{Wang2017DeepAC}. To perform the attack, we use the AC-psychic to generate a hypothetical next-state for every action given the current state, and value each of these with the assessor. We define a preference function whose output increases as the variation of potential future reward increases. In other words, we want to perturb $s_t$ when there is a large difference in the different hypothetical values $V^{\pi_T}(s_{t+1}^H)$. To craft the perturbations, we can choose $\mathcal{M}_{SRA}$ to be the proxy model of our choice (i.e. $\mathcal{M}_{S}$, $\mathcal{M}_{SR}$, or $\mathcal{M}_{SA}$). For details, see Algorithm \ref{alg:stra}.

\begin{figure*}[t]
    \centering
    \includegraphics[width=1\columnwidth]{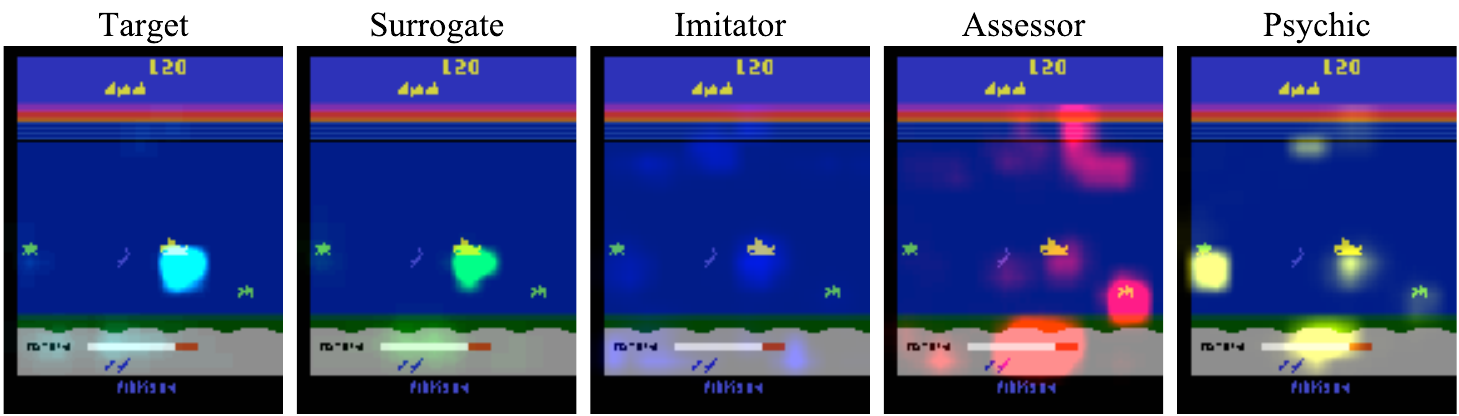}
    \caption{Comparison of features learned by target agent, surrogate agent, and proxy models on Seaquest environment using perturbation-based saliency map visualizations \cite{Greydanus2018VisualizingAU}.}
    \label{fig:seaquest_saliency}
    \vspace{-7pt}
\end{figure*}

\begin{algorithm}
\SetAlgoLined
\KwIn{Trained assessor, trained AC-psychic, trained proxy $\mathcal{M}_{\kappa}$, trained target agent $\mathcal{T}$, $\beta$}
 \For{$t$ = 1, T}{
 
  Initialize empty list $q$\;
  
  \ForEach{$a \in \mathcal{A}$}{
  
   Predict $s_{t+1}^{H}$ with \textit{AC-psychic}($s_t$, $a$)
   
   Estimate $V^H$ with \textit{assessor}($s_{t+1}^{H}$)
   
   Append $V^H$ to $q$\;
  }
  
  $c(s_t) = \max\big[\mathrm{Softmax}(q)\big] - \min\big[\mathrm{Softmax}(q)\big]$
  
  \If{$c(s_t) \ge \beta$}{
  
   Perturb $s_t$ using $\nabla_x J_{\mathcal{M}_{\kappa}}$
   
  }
  
  Feed $s_t$ to target $\mathcal{T}$ for action decision;
  
 }
 \caption{Strategically-timed snooping attack}
 \label{alg:stra}
\end{algorithm}

\section{Implementation details} \label{appendix:implementation_details}
\subsection{Agents}
We train DQN \citep{Mnih2013PlayingAW} and PPO \citep{Schulman2017ProximalPO} agents on Pong, Breakout, Space Invaders, and Seaquest games in the Atari 2600 Arcade Learning Environment \citep{Bellemare2013TheAL} via OpenAI Gym \citep{gym}. Preprocessing is primarily facilitated using the Atari wrappers in OpenAI's baselines library \citep{baselines}, which converts frames from RGB to grayscale $[0,1]$, and resizes frames to $84 \mathrm{x} 84$. To make the environment Markovian, a state is created by stacking the last four consecutive frames, making the input volume to the models of shape $4 \mathrm{x} 84 \mathrm{x} 84$. We use the PyTorch deep learning framework \citep{paszke2017automatic}, the DQN implementation described in \citep{Mnih2015HumanlevelCT}, and Ilya Kostrikov's implementation of PPO \citep{pytorchrl}.

\subsection{Proxies}
For the psychic and AC-psychic, we use a scaled-down implementation of the architecture described in \citep{Wang2017DeepAC}, as they work with the raw RGB frames and we predict the preprocessed version of the frames. The imitator architecture that we use is identical to the smaller DQN that was initially introduced in \citep{Mnih2013PlayingAW}, and apply a Softmax operation to the logits for use with the cross-entropy classification loss. Note that we intentionally use a different architecture from the target agents in the interest of strictly adhering to black-box assumptions. The assessor uses the same architecture as the imitator, but we replace the classification layer with a single output node, and train the model with the Huber regression loss \citep{Girshick2015FastR}. To train these models, we first let the trained agent collect experiences and save these into a buffer of 100,000 state/label pairs. Once we fill the buffer, we train on samples from this buffer of experiences for 30 epochs, using a batch size of 64. We repeat this process until we reach 2.5 million total training iterations. We find that this procedure significantly reduces training time compared to purely online training.

\begin{figure*}
    \centering
    \includegraphics[width=1\columnwidth]{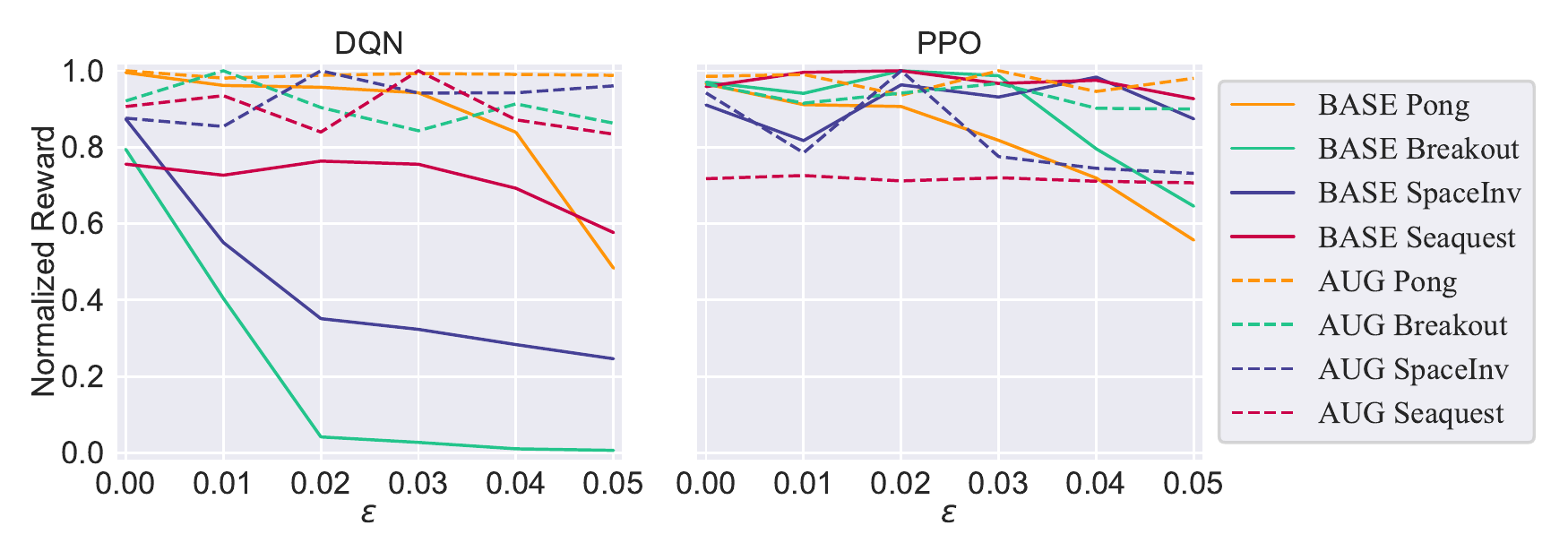}
    \caption{The effect of uniform random noise on trained agent performance. BASE agents are trained without noise and AUG agents are trained with noise. Rewards are averages over 10 episodes.}
    \label{fig:random_noise}
\end{figure*}

\section{Experiments and analysis}
Although previous work has shown that DRL agents are vulnerable to adversarial noise, to our knowledge it is not clear whether it is the \textit{adversarial} nature of the noise or simply the presence of noise that causes the agents to fail. Figure \ref{fig:random_noise} shows that small uniform random noise significantly impacts agent performance. We believe that this fragility is due to the narrow training data distribution. Atari frames do not naturally contain the noise, occlusion, or lighting variations that natural images do, which leads to a higher risk of overfitting. In the interest of proving that our attacks are effective due to transferability between proxy and agent, for the remainder of the work we only consider agents that are hardened with uniform random noise perturbations during training. Our experiments show that this produces more robust models while not significantly reducing clean data performance.

\begin{figure*}
    \centering
    \includegraphics[width=1\columnwidth]{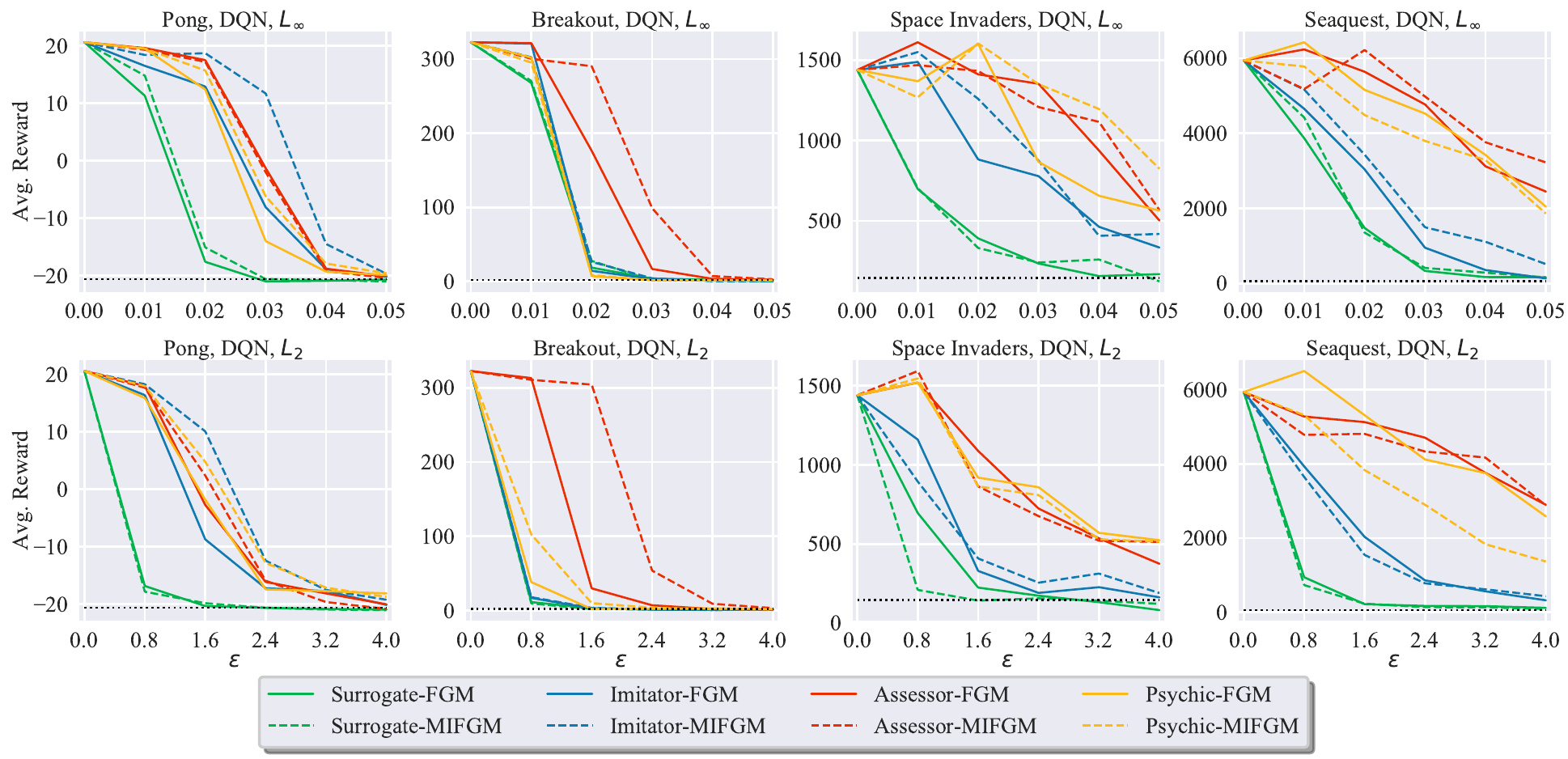}
    \caption{Performance reduction of DQN agents due to $L_\infty$ and $L_2$ bounded perturbations. The black dotted line represents a random-guess policy.}
    \label{fig:dqn_reward}
    \vspace{-10pt}
\end{figure*}

\begin{figure*}
    \centering
    \includegraphics[width=1\columnwidth]{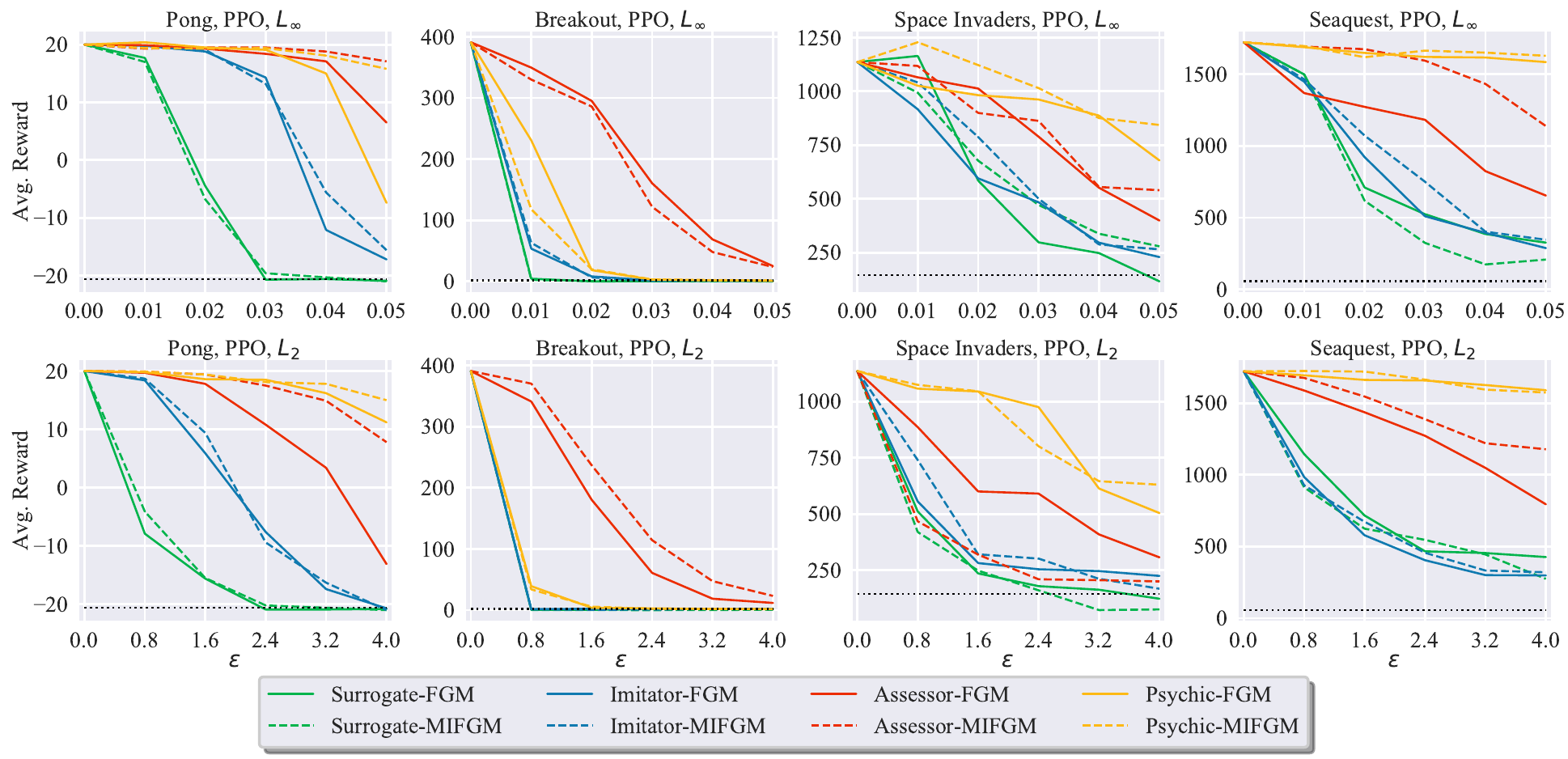}
    \caption{Performance reduction of PPO agents due to $L_\infty$ and $L_2$ bounded perturbations. The black dotted line represents a random-guess policy.}
    \label{fig:ppo_reward}
\end{figure*}

To test the effectiveness of the attacks outlined in Section \ref{sec:methodology}, we average the total reward accumulated by the target agent over 10 episodes of states perturbed by the FGM or MIFGM attacks at every time step (using the proxy gradients $\nabla_x J_{\mathcal{M}_{\kappa}}$). We repeat this on each game for five $\epsilon$ values (i.e. attack strengths). For an upper-bound baseline attack, we use the policy-transfer methodology proven to be effective in \citep{Huang2017AdversarialAO} that assumes the adversary has access to the target agent's environment and can train an identical model to the target agent. We can then use this model as a surrogate for crafting transferable adversarial examples. Figures \ref{fig:dqn_reward} and \ref{fig:ppo_reward} show the effectiveness of the transferred adversarial examples on the DQN and PPO agents, respectively.

The imitators produce the most effective attacks out of the proxy models, performing comparably to the surrogate in all games. We believe this is because these models represent the most direct proxy to the agent's policy $\pi_T$. The assessor attacks generally require a higher $\epsilon$ to reach the performance of the baseline and imitator attacks. Work by  \citet{Tramr2017TheSO} gives the insight that better performing models yield more transferable adversarial examples. We therefore hypothesize that the reason for the lower performance of the assessor attacks is because the task of predicting the exact floating point value of a given state is a difficult objective to approximate precisely given the sparsity and variance of the reward signals. The attacks from the psychics perform worse than the surrogate and imitator attacks in general. This is not surprising, as the psychics operate under the most constrained threat model. In essence, the psychic model learns a noisy encoding of what the imitator learns directly. However, note that in the Breakout environment the attacks from the psychic model roughly match the effectiveness of the surrogate and imitator attacks. We believe that this is because Breakout has a deterministic state transition function (i.e., the next frame can be exactly predicted by the current frame), unlike the other games which have state elements outside of the agent’s control (e.g., hardcoded paddle in Pong, etc.). Thus, because the psychic model's learned representation of state dynamics is a direct delineation of the agent's policy, subsequent attacks are more transferable. This result further supports our hypothesis that proxy models that more directly approximate the target agent's policy will yield more transferable attacks.

\begin{figure}
    \centering
    \includegraphics[width=.8\columnwidth]{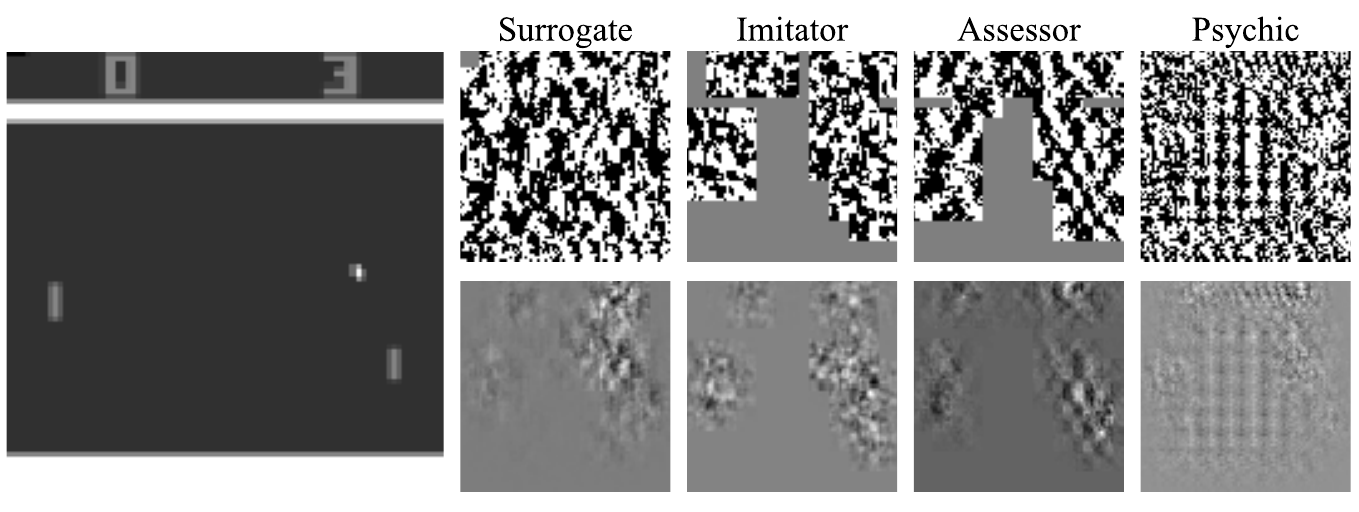}
    \caption{Noise visualization produced FGM attack by a surrogate and each proxy model for a DQN agent trained on Atari Pong environment. On the left is the current unperturbed game state. The top and bottom rows of noise are the $L_\infty$ and $L_2$ constrained perturbations, respectively. We note that the saliency patterns in the gradients of the different proxies show considerable visual similarities.}
    \label{fig:pong_adv_noise_compare}
\end{figure}

To qualitatively evaluate the features that these models learn, we create perturbation-based saliency maps \citep{Greydanus2018VisualizingAU} designed to be human interpretable visualizations of $\nabla_x J(\theta, x_i, y_i)$ on Atari games. Figure \ref{fig:seaquest_saliency} shows a representative example of what the saliency maps tell us about these models. As expected, the surrogate model's map is very similar to the target's. The imitator's salient regions tend to be the most similar to the target's out of the proxy models, which further supports our beliefs regarding attack effectiveness. The assessor tends to focus less on the agent itself (i.e. the submarine), and more on the objects around it which correspond to potential future reward. This intuition also helps to explain the inferior performance of the assessor's attacks. Finally, because the psychic model predicts the entire next-frame, it must attend to any object in the state space that moves. Because the psychic learns an approximation of the state transition dynamics, a subset of the features that it learns (e.g. the submarine, fish, divers) do correspond to features that are important to the agent's policy. Figure \ref{fig:pong_adv_noise_compare} gives a direct visualization of the gradient-based perturbations of each proxy on a Pong state. Note that the $L_2$ perturbations from the imitator and assessor proxies share considerable visual similarities to the surrogate agent's ``upper bound" perturbation. The psychic model's perturbations show notable saliency in regions of the ball and paddles, but the gradients are not as strongly localized.

\begin{figure*}
    \centering
    \includegraphics[width=1\columnwidth]{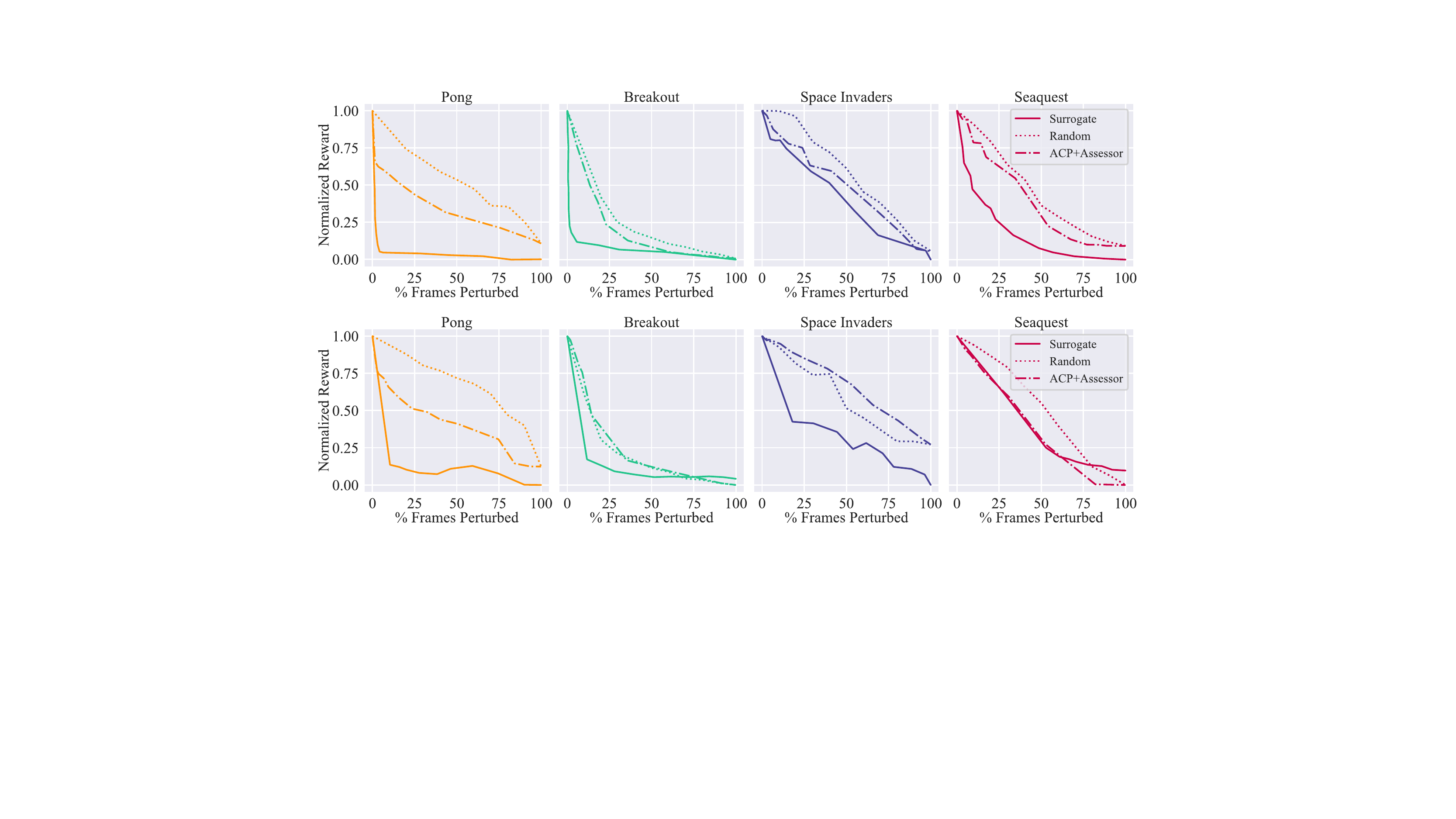}
    \caption{Effectiveness of strategically-timed attacks on agents over 30,000 time steps. Top row corresponds to DQN agents, while bottom corresponds to PPO agents. Our ACP+Assessor method and the random baseline use imitator gradients. We also compare to the technique in \citep{Lin2017TacticsOA}, which uses the stronger surrogate gradients. The FGM method is used for crafting.}
    \label{fig:combine_stra}
\end{figure*}

The results of the strategically-timed snooping attack on DQN and PPO agents can be seen in Figure \ref{fig:combine_stra}. Note that we use the FGM crafting method with imitator gradients in this experiment. We compare this performance to (1) the black-box variant of the original strategically-timed attack described in \citep{Lin2017TacticsOA} using FGM with surrogate gradients, and (2) an attack which perturbs at random time steps with FGM and imitator gradients. The strategically-timed attacks will typically be more effective on the paddle-based games such as Pong, as there is a clear distinction between time steps that are critical for good performance (i.e. as the ball approaches the paddle) and time steps when the actions of the agent do not matter. As expected, our method does not perform as well as the surrogate that was trained to assign credit to certain states and actions. Our method does show promise, however, as it performs markedly better than the random strategy without requiring any interaction with the environment. We believe that our method's effectiveness is bottlenecked by the high-variance assessor predictions, and reducing the variance of the reward signal learned would lead to better performance.

An unexpected result of our experiments is the unimpressive performance of MIFGM compared to vanilla FGM. To understand why, we consider that iterative attacks like MIFGM create a more custom perturbation for the model that it is crafted by. This introduces the possibility that the resulting adversarial examples are overfit to a local maxima of the white-box model. In the case of the the proxy models, we posit that because of the significant differences in the loss functions used by the agents and the proxies, adversarial examples that are more tailored to a specific objective will not perform as well. In the case of the baseline surrogate agents that \textit{do} use the same objectives, the high variance of the RL data generation process is more likely to yield models with differently-shaped decision surfaces compared to supervised classifiers that draw i.i.d. samples. Again, this renders these more tailored perturbations less effective for transfer-based attacks. 

Finally, we propose that our work offers a unique prospective on adversarial example transferability. Previous work has shown that adversarial examples transfer between different classifiers, etc. This phenomena has been justified by decision surface similarity between the classifiers \citep{Liu2017DelvingIT, Tramr2017TheSO, Papernot2016TransferabilityIM}. However, to our knowledge, we are among the first to demonstrate that adversarial examples transfer across models that learn with very different optimization objectives as long as the tasks are related such that the gradient of the loss with respect to the input has a similar shape. For example, although the loss, output dimensionality, and decision surface of the psychic proxy is radically different from the target model, adversarial examples generated by the psychic also fool the agent. This implies that an adversary can effectively attack agents \textit{even if they have no knowledge of what the target system is trying to do}. These findings alone force us to reconsider what makes adversarial examples transfer between models. We believe that our snooping threat models and proxies offer a unique and effective test environment for further work on the study of transferability of adversarial examples between models with fundamentally different but \textit{related} objectives.

\section{Conclusions}
It is often infeasible for an adversary to have access to a target agent's parameters or train a DRL surrogate agent on a proprietary environment. However, we show that an adversary only needs the ability to intercept the states and optionally snoop on the action and reward signals to launch powerful adversarial attacks on a black-box agent. Further, we show that adversaries can use trained proxy models to make their attacks sparse in time, rendering them more difficult to detect. We believe these to be important security vulnerabilities to consider and address before deploying these models to systems that we trust.

\bibliography{lit}  

\begin{thebibliography}{40}
\providecommand{\natexlab}[1]{#1}
\providecommand{\url}[1]{\texttt{#1}}
\expandafter\ifx\csname urlstyle\endcsname\relax
  \providecommand{\doi}[1]{doi: #1}\else
  \providecommand{\doi}{doi: \begingroup \urlstyle{rm}\Url}\fi

\bibitem[Behzadan and Munir(2017)]{Behzadan2017VulnerabilityOD}
Vahid Behzadan and Arslan Munir.
\newblock Vulnerability of deep reinforcement learning to policy induction
  attacks.
\newblock In \emph{MLDM}, 2017.

\bibitem[Bellemare et~al.(2013)Bellemare, Naddaf, Veness, and
  Bowling]{Bellemare2013TheAL}
Marc~G. Bellemare, Yavar Naddaf, Joel Veness, and Michael~H. Bowling.
\newblock The arcade learning environment: An evaluation platform for general
  agents.
\newblock \emph{J. Artif. Intell. Res.}, 47:\penalty0 253--279, 2013.

\bibitem[Bellman(1966)]{Bellman1966DynamicP}
Richard Bellman.
\newblock Dynamic programming.
\newblock \emph{Science}, 153 3731:\penalty0 34--7, 1966.

\bibitem[Bojarski et~al.(2016)Bojarski, Testa, Dworakowski, Firner, Flepp,
  Goyal, Jackel, Monfort, Muller, Zhang, Zhang, Zhao, and
  Zieba]{Bojarski2016EndTE}
Mariusz Bojarski, Davide~Del Testa, Daniel Dworakowski, Bernhard Firner, Beat
  Flepp, Prasoon Goyal, Lawrence~D. Jackel, Miguel~Pozuelo Monfort, Urs Muller,
  Jiakai Zhang, Xin Zhang, Junbo~Jake Zhao, and Karol Zieba.
\newblock End to end learning for self-driving cars.
\newblock \emph{CoRR}, abs/1604.07316, 2016.

\bibitem[Brockman et~al.(2016)Brockman, Cheung, Pettersson, Schneider,
  Schulman, Tang, and Zaremba]{gym}
Greg Brockman, Vicki Cheung, Ludwig Pettersson, Jonas Schneider, John Schulman,
  Jie Tang, and Wojciech Zaremba.
\newblock Openai gym, 2016.

\bibitem[Carlini and Wagner(2017)]{Carlini2017TowardsET}
Nicholas Carlini and David~A. Wagner.
\newblock Towards evaluating the robustness of neural networks.
\newblock \emph{2017 IEEE Symposium on Security and Privacy (SP)}, pages
  39--57, 2017.

\bibitem[Dhariwal et~al.(2017)Dhariwal, Hesse, Klimov, Nichol, Plappert,
  Radford, Schulman, Sidor, Wu, and Zhokhov]{baselines}
Prafulla Dhariwal, Christopher Hesse, Oleg Klimov, Alex Nichol, Matthias
  Plappert, Alec Radford, John Schulman, Szymon Sidor, Yuhuai Wu, and Peter
  Zhokhov.
\newblock Openai baselines.
\newblock \url{https://github.com/openai/baselines}, 2017.

\bibitem[Dong et~al.(2018)Dong, Liao, Pang, Su, Zhu, Hu, and
  Li]{Dong2018BoostingAA}
Yinpeng Dong, Fangzhou Liao, Tianyu Pang, Hang Su, Jun Zhu, Xiaolin Hu, and
  Jianguo Li.
\newblock Boosting adversarial attacks with momentum.
\newblock In \emph{CVPR}, 2018.

\bibitem[Girshick(2015)]{Girshick2015FastR}
Ross~B. Girshick.
\newblock Fast r-cnn.
\newblock \emph{2015 IEEE International Conference on Computer Vision (ICCV)},
  pages 1440--1448, 2015.

\bibitem[Goodfellow et~al.(2015)Goodfellow, Shlens, and
  Szegedy]{Goodfellow2015ExplainingAH}
Ian~J. Goodfellow, Jonathon Shlens, and Christian Szegedy.
\newblock Explaining and harnessing adversarial examples.
\newblock \emph{CoRR}, abs/1412.6572, 2015.

\bibitem[Greydanus et~al.(2018)Greydanus, Koul, Dodge, and
  Fern]{Greydanus2018VisualizingAU}
Sam Greydanus, Anurag Koul, Jonathan Dodge, and Alan Fern.
\newblock Visualizing and understanding atari agents.
\newblock In \emph{ICML}, 2018.

\bibitem[Huang et~al.(2017)Huang, Papernot, Goodfellow, Duan, and
  Abbeel]{Huang2017AdversarialAO}
Sandy~H. Huang, Nicolas Papernot, Ian~J. Goodfellow, Yan Duan, and Pieter
  Abbeel.
\newblock Adversarial attacks on neural network policies.
\newblock \emph{CoRR}, abs/1702.02284, 2017.

\bibitem[Inkawhich et~al.(2019)Inkawhich, Wen, Li, and
  Chen]{Inkawhich2019FeatureSpaceP}
Nathan Inkawhich, Wei Wen, Hai Li, and Yiran Chen.
\newblock Feature space perturbations yield more transferable adversarial
  examples.
\newblock In \emph{CVPR}, 2019.

\bibitem[Kostrikov(2018)]{pytorchrl}
Ilya Kostrikov.
\newblock Pytorch implementations of reinforcement learning algorithms.
\newblock \url{https://github.com/ikostrikov/pytorch-a2c-ppo-acktr-gail}, 2018.

\bibitem[Leibfried et~al.(2017)Leibfried, Kushman, and
  Hofmann]{Leibfried2017ADL}
Felix Leibfried, Nate Kushman, and Katja Hofmann.
\newblock A deep learning approach for joint video frame and reward prediction
  in atari games.
\newblock \emph{CoRR}, abs/1611.07078, 2017.

\bibitem[Levine et~al.(2016)Levine, Finn, Darrell, and
  Abbeel]{Levine2016EndtoEndTO}
Sergey Levine, Chelsea Finn, Trevor Darrell, and Pieter Abbeel.
\newblock End-to-end training of deep visuomotor policies.
\newblock \emph{Journal of Machine Learning Research}, 17:\penalty0
  39:1--39:40, 2016.

\bibitem[Lillicrap et~al.(2016)Lillicrap, Hunt, Pritzel, Heess, Erez, Tassa,
  Silver, and Wierstra]{Lillicrap2016ContinuousCW}
Timothy~P. Lillicrap, Jonathan~J. Hunt, Alexander Pritzel, Nicolas Heess, Tom
  Erez, Yuval Tassa, David Silver, and Daan Wierstra.
\newblock Continuous control with deep reinforcement learning.
\newblock \emph{CoRR}, abs/1509.02971, 2016.

\bibitem[Lin et~al.(2017)Lin, Hong, Liao, Shih, Liu, and Sun]{Lin2017TacticsOA}
Yen-Chen Lin, Zhang-Wei Hong, Yuan-Hong Liao, Meng-Li Shih, Ming-Yu Liu, and
  Min Sun.
\newblock Tactics of adversarial attack on deep reinforcement learning agents.
\newblock In \emph{IJCAI}, 2017.

\bibitem[Liu et~al.(2017)Liu, Chen, Liu, and Song]{Liu2017DelvingIT}
Yanpei Liu, Xinyun Chen, Chang Liu, and Dawn~Xiaodong Song.
\newblock Delving into transferable adversarial examples and black-box attacks.
\newblock \emph{CoRR}, abs/1611.02770, 2017.

\bibitem[Mnih et~al.(2013)Mnih, Kavukcuoglu, Silver, Graves, Antonoglou,
  Wierstra, and Riedmiller]{Mnih2013PlayingAW}
Volodymyr Mnih, Koray Kavukcuoglu, David Silver, Alex Graves, Ioannis
  Antonoglou, Daan Wierstra, and Martin~A. Riedmiller.
\newblock Playing atari with deep reinforcement learning.
\newblock \emph{CoRR}, abs/1312.5602, 2013.

\bibitem[Mnih et~al.(2015)Mnih, Kavukcuoglu, Silver, Rusu, Veness, Bellemare,
  Graves, Riedmiller, Fidjeland, Ostrovski, Petersen, Beattie, Sadik,
  Antonoglou, King, Kumaran, Wierstra, Legg, and
  Hassabis]{Mnih2015HumanlevelCT}
Volodymyr Mnih, Koray Kavukcuoglu, David Silver, Andrei~A. Rusu, Joel Veness,
  Marc~G. Bellemare, Alex Graves, Martin~A. Riedmiller, Andreas Fidjeland,
  Georg Ostrovski, Stig Petersen, Charles Beattie, Amir Sadik, Ioannis
  Antonoglou, Helen King, Dharshan Kumaran, Daan Wierstra, Shane Legg, and
  Demis Hassabis.
\newblock Human-level control through deep reinforcement learning.
\newblock \emph{Nature}, 518:\penalty0 529--533, 2015.

\bibitem[Mnih et~al.(2016)Mnih, Badia, Mirza, Graves, Lillicrap, Harley,
  Silver, and Kavukcuoglu]{Mnih2016AsynchronousMF}
Volodymyr Mnih, Adri{\`a}~Puigdom{\`e}nech Badia, Mehdi Mirza, Alex Graves,
  Timothy~P. Lillicrap, Tim Harley, David Silver, and Koray Kavukcuoglu.
\newblock Asynchronous methods for deep reinforcement learning.
\newblock In \emph{ICML}, 2016.

\bibitem[Oh et~al.(2015)Oh, Guo, Lee, Lewis, and
  Singh]{Oh2015ActionConditionalVP}
Junhyuk Oh, Xiaoxiao Guo, Honglak Lee, Richard~L. Lewis, and Satinder~P. Singh.
\newblock Action-conditional video prediction using deep networks in atari
  games.
\newblock In \emph{NIPS}, 2015.

\bibitem[Papernot et~al.(2016{\natexlab{a}})Papernot, McDaniel, and
  Goodfellow]{Papernot2016TransferabilityIM}
Nicolas Papernot, Patrick~D. McDaniel, and Ian~J. Goodfellow.
\newblock Transferability in machine learning: from phenomena to black-box
  attacks using adversarial samples.
\newblock \emph{CoRR}, abs/1605.07277, 2016{\natexlab{a}}.

\bibitem[Papernot et~al.(2016{\natexlab{b}})Papernot, McDaniel, Jha,
  Fredrikson, Celik, and Swami]{Papernot2016TheLO}
Nicolas Papernot, Patrick~D. McDaniel, Somesh Jha, Matt Fredrikson, Z.~Berkay
  Celik, and Ananthram Swami.
\newblock The limitations of deep learning in adversarial settings.
\newblock \emph{2016 IEEE European Symposium on Security and Privacy
  (EuroS\&P)}, pages 372--387, 2016{\natexlab{b}}.

\bibitem[Papernot et~al.(2017)Papernot, McDaniel, Goodfellow, Jha, Celik, and
  Swami]{Papernot2017PracticalBA}
Nicolas Papernot, Patrick~D. McDaniel, Ian~J. Goodfellow, Somesh Jha, Z.~Berkay
  Celik, and Ananthram Swami.
\newblock Practical black-box attacks against machine learning.
\newblock In \emph{AsiaCCS}, 2017.

\bibitem[Paszke et~al.(2017)Paszke, Gross, Chintala, Chanan, Yang, DeVito, Lin,
  Desmaison, Antiga, and Lerer]{paszke2017automatic}
Adam Paszke, Sam Gross, Soumith Chintala, Gregory Chanan, Edward Yang, Zachary
  DeVito, Zeming Lin, Alban Desmaison, Luca Antiga, and Adam Lerer.
\newblock Automatic differentiation in pytorch.
\newblock 2017.

\bibitem[Pathak et~al.(2018)Pathak, Mahmoudieh, Luo, Agrawal, Chen, Shentu,
  Shelhamer, Malik, Efros, and Darrell]{Pathak2018ZeroShotVI}
Deepak Pathak, Parsa Mahmoudieh, Guanghao Luo, Pulkit Agrawal, Dian Chen, Yide
  Shentu, Evan Shelhamer, Jitendra Malik, Alexei~A. Efros, and Trevor Darrell.
\newblock Zero-shot visual imitation.
\newblock \emph{2018 IEEE/CVF Conference on Computer Vision and Pattern
  Recognition Workshops (CVPRW)}, pages 2131--21313, 2018.

\bibitem[Pattanaik et~al.(2018)Pattanaik, Tang, Liu, Bommannan, and
  Chowdhary]{Pattanaik2018RobustDR}
Anay Pattanaik, Zhenyi Tang, Shuijing Liu, Gautham Bommannan, and Girish
  Chowdhary.
\newblock Robust deep reinforcement learning with adversarial attacks.
\newblock In \emph{AAMAS}, 2018.

\bibitem[Pinto et~al.(2018)Pinto, Andrychowicz, Welinder, Zaremba, and
  Abbeel]{Pinto2018AsymmetricAC}
Lerrel Pinto, Marcin Andrychowicz, Peter Welinder, Wojciech Zaremba, and Pieter
  Abbeel.
\newblock Asymmetric actor critic for image-based robot learning.
\newblock \emph{CoRR}, abs/1710.06542, 2018.

\bibitem[Sallab et~al.(2017)Sallab, Abdou, Perot, and
  Yogamani]{Sallab2017DeepRL}
Ahmad~El Sallab, Mohammed Abdou, Etienne Perot, and Senthil Yogamani.
\newblock Deep reinforcement learning framework for autonomous driving.
\newblock \emph{CoRR}, abs/1704.02532, 2017.

\bibitem[Schulman et~al.(2015)Schulman, Levine, Moritz, Jordan, and
  Abbeel]{Schulman2015TrustRP}
John Schulman, Sergey Levine, Philipp Moritz, Michael~I. Jordan, and Pieter
  Abbeel.
\newblock Trust region policy optimization.
\newblock In \emph{ICML}, 2015.

\bibitem[Schulman et~al.(2017)Schulman, Wolski, Dhariwal, Radford, and
  Klimov]{Schulman2017ProximalPO}
John Schulman, Filip Wolski, Prafulla Dhariwal, Alec Radford, and Oleg Klimov.
\newblock Proximal policy optimization algorithms.
\newblock \emph{CoRR}, abs/1707.06347, 2017.

\bibitem[Silver et~al.(2014)Silver, Lever, Heess, Degris, Wierstra, and
  Riedmiller]{Silver2014DeterministicPG}
David Silver, Guy Lever, Nicolas Heess, Thomas Degris, Daan Wierstra, and
  Martin~A. Riedmiller.
\newblock Deterministic policy gradient algorithms.
\newblock In \emph{ICML}, 2014.

\bibitem[Silver et~al.(2016)Silver, Huang, Maddison, Guez, Sifre, van~den
  Driessche, Schrittwieser, Antonoglou, Panneershelvam, Lanctot, Dieleman,
  Grewe, Nham, Kalchbrenner, Sutskever, Lillicrap, Leach, Kavukcuoglu, Graepel,
  and Hassabis]{Silver2016MasteringTG}
David Silver, Aja Huang, Chris~J. Maddison, Arthur Guez, Laurent Sifre, George
  van~den Driessche, Julian Schrittwieser, Ioannis Antonoglou, Vedavyas
  Panneershelvam, Marc Lanctot, Sander Dieleman, Dominik Grewe, John Nham, Nal
  Kalchbrenner, Ilya Sutskever, Timothy~P. Lillicrap, Madeleine Leach, Koray
  Kavukcuoglu, Thore Graepel, and Demis Hassabis.
\newblock Mastering the game of go with deep neural networks and tree search.
\newblock \emph{Nature}, 529:\penalty0 484--489, 2016.

\bibitem[Sutton and Barto(1988)]{Sutton1988ReinforcementLA}
Richard~S. Sutton and Andrew~G. Barto.
\newblock Reinforcement learning: An introduction.
\newblock \emph{IEEE Transactions on Neural Networks}, 16:\penalty0 285--286,
  1988.

\bibitem[Szegedy et~al.(2014)Szegedy, Zaremba, Sutskever, Bruna, Erhan,
  Goodfellow, and Fergus]{Szegedy2014IntriguingPO}
Christian Szegedy, Wojciech Zaremba, Ilya Sutskever, Joan Bruna, Dumitru Erhan,
  Ian~J. Goodfellow, and Rob Fergus.
\newblock Intriguing properties of neural networks.
\newblock \emph{CoRR}, abs/1312.6199, 2014.

\bibitem[Tram{\`e}r et~al.(2017)Tram{\`e}r, Papernot, Goodfellow, Boneh, and
  McDaniel]{Tramr2017TheSO}
Florian Tram{\`e}r, Nicolas Papernot, Ian~J. Goodfellow, Dan Boneh, and
  Patrick~D. McDaniel.
\newblock The space of transferable adversarial examples.
\newblock \emph{CoRR}, abs/1704.03453, 2017.

\bibitem[Wang(2017)]{Wang2017DeepAC}
Elias Wang.
\newblock Deep action conditional neural network for frame prediction in atari
  games.
\newblock 2017.

\bibitem[Watkins and Dayan(1992)]{Watkins1992TechnicalNQ}
Christopher J. C.~H. Watkins and Peter Dayan.
\newblock Technical note: Q-learning.
\newblock \emph{Machine Learning}, 8:\penalty0 279--292, 1992.

\end{thebibliography}

\end{document}